\documentclass[runningheads]{llncs}
\usepackage{graphicx}

\usepackage{tikz}
\usepackage{comment}
\usepackage{amsmath,amssymb} 
\usepackage{color}
\usepackage[pagebackref,breaklinks,colorlinks]{hyperref}
\usepackage{gensymb}
\usepackage{mathrsfs}
\usepackage{booktabs}
\usepackage{array}
\usepackage{bm}
\usepackage{lipsum}
\usepackage{makecell}
\usepackage{microtype}
\usepackage{cite}
\usepackage{multicol} 
\usepackage{caption} 
\usepackage{subcaption} 
\usepackage{float}
\usepackage{pifont}
\usepackage{multirow}

\usepackage{newfloat}
\usepackage{listings}

\usepackage[accsupp]{axessibility}  

\begin{document}
\pagestyle{headings}
\mainmatter
\def\ECCVSubNumber{5599}  

\title{D2C-SR: A Divergence to Convergence Approach for Real-World Image Super-Resolution} 

\titlerunning{D2C-SR}
%
\author{Youwei Li\,$^1$\quad
Haibin Huang\,$^2$\quad
Lanpeng Jia\,$^3$\quad
Haoqiang Fan\,$^1$\quad \\
Shuaicheng Liu\,$^{4,1,\dagger}$}

\authorrunning{Y. Li et al.}

\institute{$^1$\,Megvii Technology, Beijing, China \quad$^2$\,Kuaishou Technology, Beijing, China\\
$^3$\,Great Wall Motor Company Limited, Baoding, China \\
$^4$\,University of Electronic Science and Technology of China, Chengdu, China
\email{\{liyouwei, fhq\}@megvii.com} 
\email{jialanpeng@gwm.cn} 
\email{jackiehuanghaibin@gmail.com} 
\email{liushuaicheng@uestc.edu.cn}\\
$^\dagger$Corresponding Author
}
\maketitle

\begin{abstract}
In this paper, we present D2C-SR, a novel framework for the task of real-world image super-resolution. As an ill-posed problem, the key challenge in super-resolution related tasks is there can be multiple predictions for a given low-resolution input. Most classical deep learning based approaches ignored the fundamental fact and lack explicit modeling of the underlying high-frequency distribution which leads to blurred results. Recently, some methods of GAN-based or learning super-resolution space can generate simulated textures but do not promise the accuracy of the textures which have low quantitative performance. Rethinking both, we learn the distribution of underlying high-frequency details in a discrete form and propose a two-stage pipeline: divergence stage to convergence stage. At divergence stage, we propose a tree-based structure deep network as our divergence backbone. Divergence loss is proposed to encourage the generated results from the tree-based network to diverge into possible high-frequency representations, which is our way of discretely modeling the underlying high-frequency distribution. At convergence stage, we assign spatial weights to fuse these divergent predictions to obtain the final output with more accurate details. Our approach provides a convenient end-to-end manner to inference. We conduct evaluations on several real-world benchmarks, including a new proposed D2CRealSR dataset with x8 scaling factor. Our experiments demonstrate that D2C-SR achieves better accuracy and visual improvements against state-of-the-art methods, with a significantly less parameters number and our D2C structure can also be applied as a generalized structure to some other methods to obtain improvement. Our codes and dataset are available at \href{https://github.com/megvii-research/D2C-SR}{https://github.com/megvii-research/D2C-SR}. 

\end{abstract}

\section{Introduction}
\label{sec:introduction}

Super-resolution (SR) is one of the fundamental problems in computer vision with its applications for several important image processing tasks. Despite decades of development in SR technologies, it remains challenging to recover high-quality and accurate details from low-resolution (LR) inputs due to its ill-posed nature.

In fact, given a LR image, there exist infinitely possible high-resolution (HR) predictions by adding different high-frequency information. This is the key challenge for the design of learning-based SR methods. Recently, deep neural network~(DNN) based methods~\cite{simonyan2014very,lim2017enhanced,zhang2018image,michelini2019multigrid,song2020efficient} have achieved tremendous success and outperformed most classical methods based on sparse coding~\cite{yang2008image,dai2015jointly} or local liner regression~\cite{yang2013fast,timofte2014a+}. However, these early deep learning methods only rely on $L_1$ or $L_2$ reconstruction loss, which have a relatively high quantitative performance but can not reproduce rich texture details due to ill-posed nature. Later on, conditional GAN-based methods~\cite{goodfellow2014generative} are adapted into SR tasks, which provide methods for learning the distribution in a data-driven way and led to generate richer simulated textures. However, these methods still suffered from mode collapse and tend to generate implausible or inaccurate texture details. Thus, the quantitative performance of GAN-based methods are not satisfactory. After realizing the importance of modeling the underlying high-frequency distribution explicitly, some methods of learning SR space appear. SRFlow~\cite{lugmayr2020srflow} introduced a normalizing flow-based method which tried to address the mode collapse problem and make a continuous Gaussian distribution to model all the high-frequency. Therefore, although SRFlow can sample to get a certain prediction, it does not promise the most accurate prediction. Furthermore, when adapting more complicated distribution, like large amount of real-world scenes, it would significantly increase the hardness of convergence during training. Some works ~\cite{kim2016deeply,wang2017ensemble,xiong2018gradient,pan2020real} introduced ensemble-based SR approaches that trained a same network from different initialization or different down-sampling methods and merged the outputs into the SR results. Due to the lack of explicit modeling of the SR space, the different results produced by these methods are very dependent on the randomness of the initialization or artificially rules, and are therefore unstable and has a risk of degradation. In other words, they lack an explicit and stable way to make the results divergent. However, these works still give us meaningful inspiration. In this paper, some discussions are made based on this.

In this paper, we present D2C-SR, a novel divergence to convergence framework for real-world SR, by rethinking the ill-posed problem and the SR space approaches. D2C-SR follows the idea of explicitly learning the underlying distribution of high-frequency details. But unlike conditional GAN-based methods or SRFlow~\cite{lugmayr2020srflow}, our D2C-SR model the high-frequency details distribution using a discrete manner. Our key insight is that: most classical SR methods use a single output to fit all high-frequency details directly and `regression to mean', therefore only obtains average and blurred outputs. While SRFlow~\cite{lugmayr2020srflow} uses an continuous Gaussian distribution to fit all high-frequency details, but the fit is very difficult and does not guarantee stable and accurate high-frequency details. The GAN-based approaches have the same problem. Therefore, we adopt a trade-off approach by using a finite discrete distribution to fit the high-frequency details, thus ensuring a stable, controllable and accurate output relatively. 

Specifically, at divergence stage, we first propose a tree-based structure network, where end branches are designed to learn random possible high-frequency details separately. To stabilize the divergence process, we also propose the divergence loss which can explicitly allows for divergence of the results produced by different branches. Meanwhile, we use the construction loss to ensure consistency with HR during the entire process. Due to there are multiple local minima in the network optimization process, we expect different branches to have access to different local minima through the appropriate strength of these restraints. Theoretically, if there are multiple branches and fit to the full range of high-frequency possibilities, the best HR image should be the fusion of the high-frequency details exactly. Hence, at convergence stage, we assign spatial weights to combine the divergent predictions to produce a more accurate result. 

To fully evaluate the efficiency and generality of D2C-SR for real-world SISR, we conduct experiments on several benchmarks, including RealSR\cite{cai2019toward}, DRealSR\cite{wei2020component} and our new benchmark, D2CRealSR, with x8 upscaling factor. Experimental results show that our D2C-SR can achieve state-of-the-art performance and visual improvements with less parameters. In sum, the main contributions are as follows:
\begin{itemize}

\item We present D2C-SR, a novel framework with divergence and convergence stage for real-world SR. D2C-SR explicitly model the underlying distribution of high-frequency details in a discrete manner and provide a convenient end-to-end inference. For stable divergence, we also propose the divergence loss to generated multiple results with divergent high-frequency representations.

\item A new real-world SR benchmark (D2CRealSR), which has a larger scaling factor (x8) compared to existing real-world SR benchmarks.

\item D2C-SR sets up new state-of-the-art performance on many popular real-world SR benchmarks, including our new proposed D2CRealSR benchmark and, it can provide compatible performance with significantly less parameter number.

\item Our D2C structure can also be applied as a generalized structure to some other methods to obtain improvement.

\end{itemize}

\section{Related Work}

\subsection{DNN-based SISR}
Single Image Super-Resolution (SISR) is a long standing research topic due to its importance and ill-posed nature. Traditional learning-based methods adopts sparse coding~\cite{dai2015jointly,sun2012super,yang2008image} or local linear regression~\cite{timofte2014a+,timofte2013anchored,yang2013fast}. Deep learning (DL)-based methods have achieved dramatic advantages against conventional methods for SISR~\cite{wang2020deep,anwar2020deep}. It is first proposed by SRCNN~\cite{simonyan2014very} that employs a relatively shallow network and adopts the bicubic degradation for HR and LR pairs. Following which, various SISR approaches have been proposed, such as VDSR that adopts very deep network~\cite{simonyan2014very}; EDSR that modifies the ResNet for enhancement~\cite{lim2017enhanced}; ESPCN that uses efficient sub-pixel CNN~\cite{shi2016real}; CDC that divides images into multiple regions~\cite{wei2020component}, and VGG loss~\cite{simonyan2014very}, GAN loss~\cite{goodfellow2014generative} that improve the perceptual visual quality~\cite{ledig2017photo,sajjadi2017enhancenet,wang2018esrgan}. Ensemble-based methods ~\cite{kim2016deeply,wang2017ensemble,xiong2018gradient,pan2020real} train a same network from different initialization or different training pair generation but it lacks explicit modeling of SR space. In this work, we propose a tree-based network structure, for the purpose of multi-mode learning.

\begin{figure}[h]

\begin{center}
\includegraphics[width=0.7\linewidth]{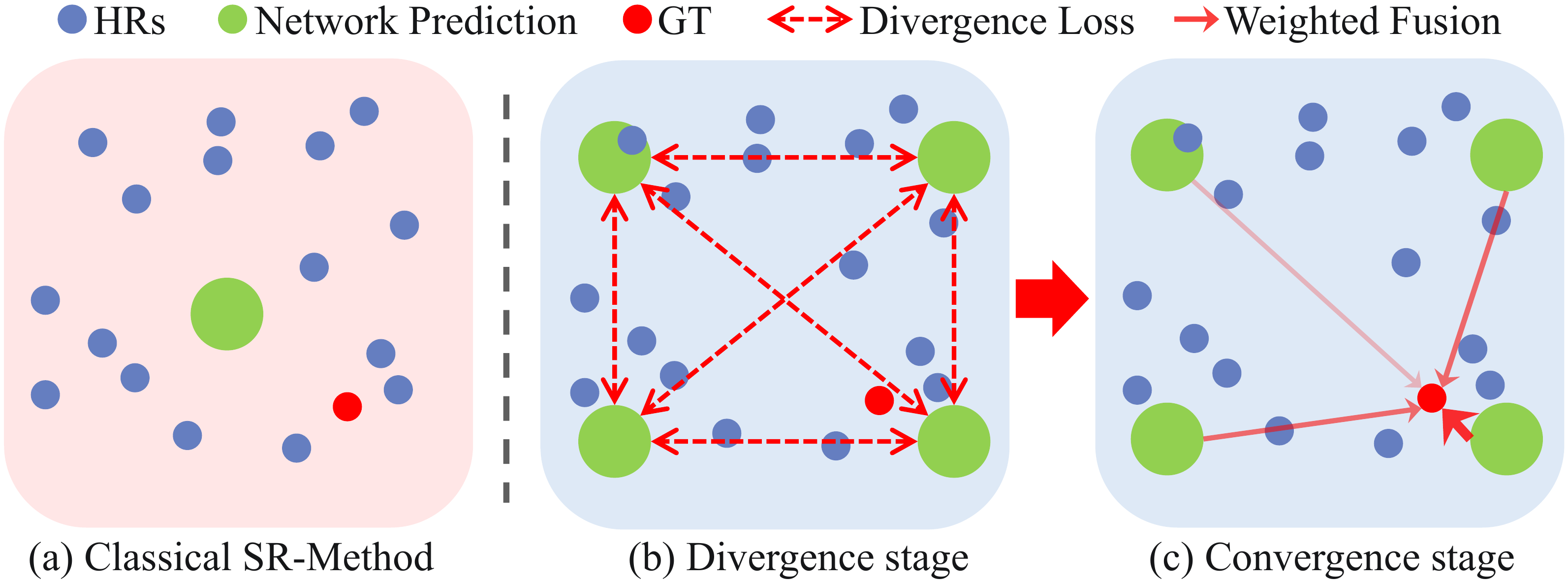}
\end{center}

  \caption{
  Difference between our method and the classical methods in handling ill-posed problems. Blue and red dots in the box indicate possible HR texture details and the ideal details for a settled LR region. Classical methods only find the single average prediction. Our method can produce multiple divergent predictions at divergence stage and fuse to obtain more accurate predictions at convergence stage.}
\label{fig:method}

\end{figure}

\subsection{Learning Super-Resolution Space}

Single image super-resolution is ill-posed, where infinitely many high-resolution images can be downsampled to the same low-resolution images. Therefore learning a deterministic mapping may not be optimal. The problem can be converted to stochastic mapping, where multiple plausible high-resolution images are predicted given a single low-resolution input~\cite{bahat2020explorable}. DeepSEE incorporates semantic maps for explorative facial super-resolution~\cite{buhler2020deepsee}. SRFlow adopts normalizing flow to learn the high-frequency distribution of facial datasets by using an continuous Gaussian distribution~\cite{lugmayr2020srflow}. Unlike this, our method model the high-frequency distribution using a discrete manner. Thus the prediction of our method has better texture consistency with HR and the details are more accurate.

\subsection{Real-World Super-Resolution and Datasets}

Bicubic downsampling datasets are widely used but the gap between the simulation and the real degradation limits the performance in the real-world application~\cite{gu2019blind}. Therefore, some works explore degradation of real scenarios by collecting real-world datasets, e.g., City100~\cite{chen2019camera}, RealSR~\cite{cai2019toward}, SR-RAW~\cite{zhang2019zoom} and DRealSR~\cite{wei2020component}. We propose the D2CRealSR dataset with larger scaling factor (x8) compared to the above datasets. Real-world degradation loses more details compared to simulated degradation~\cite{chen2019camera}, and the reduction of the prior makes recovery more difficult. Our method can handle these problems better than others.

\section{Method}
\label{sec Method}

\subsection{Overview}

\begin{figure}[]
\begin{center}
\includegraphics[width=1.0\linewidth]{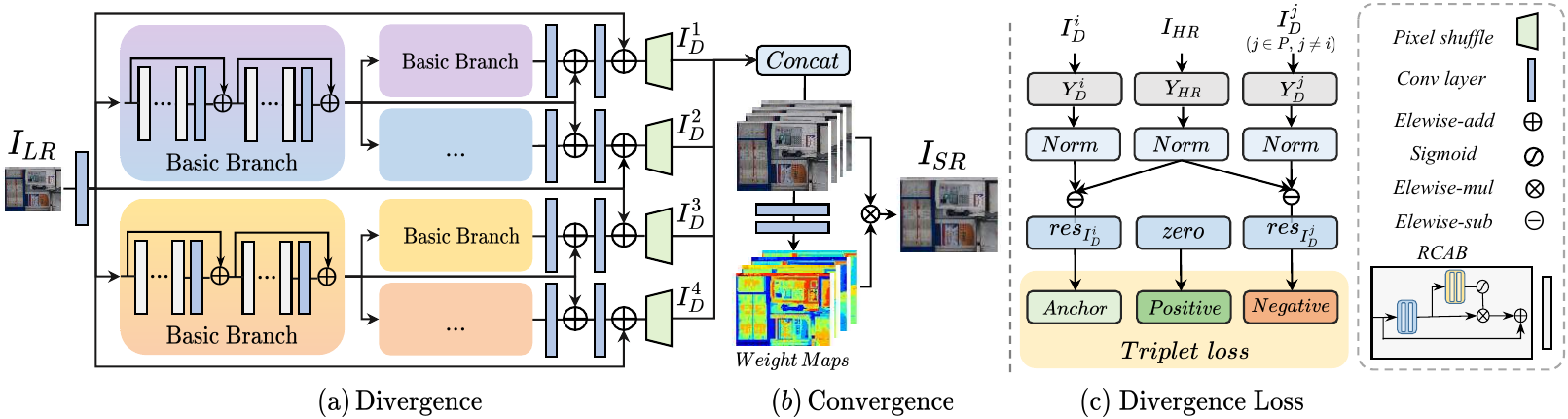}
\end{center}

   \caption{Two stages in D2C architecture: (a) Divergence stage, (b) Convergence stage. Divergence network with tree-based structure outputs multi-predictions $I_{D}^i$ with different high-frequency recovery. Convergence network obtains more accurate result by weighted combining divergence results. (c) Divergence loss.
}
\label{fig:pipeline}

\end{figure}

As shown in Fig.~\ref{fig:pipeline}, D2C-SR consists of two sub-networks: a divergence network to learn the possible predictions and a convergence network to fuse the predictions to final result. In this section, we first describe divergence network in Sec.~\ref{sec Divergence Network}. Then, we introduce convergence network in Sec.~\ref{sec Convergence Network}. The network training strategy are described in Sec.~\ref{sec Training Strategy}. To facilitate the explanation, we also present in Fig.~\ref{fig:method}.

\subsection{Divergence Network}

\label{sec Divergence Network}
Divergence network produce multiple divergent predictions to address the ill-posed nature. We built a tree-based structure network to get the desired predictions. There are four main modules in the divergence network: shallow feature extraction, basic branch module, multi-branch deep residual structure and upscale module. We define the depth of the tree network is $L$, and each branch has $C$ sub-branches. Each branch and its sub-branches are composed of a basic branch module and non-weight shared. Further, the basic branch module contains $G$ residual groups, and each residual group contains $B$ residual channel attention blocks, which refer to RCAN~\cite{zhang2018image}. The divergence network first extracts the shallow features from the input LR image by the shallow feature extraction module, which is a simple Conv layer. Then, the input features are each fed into their own branch and sub-branches modules. The feature outputs of the end branches go through the multi-branch deep residual structure, then upscaled by the upscale module. So far, the divergence network generate $P$ divergent predictions. These predictions are expressed as $I_{D}=\mathcal{F}(I_{LR};\Theta_{D})$, where $\Theta_{D}$ is the divergence network parameter and $I_{LR}$ is the LR image. $I_{D}^{i}$ denotes $i$-th prediction of divergence network.

\subsubsection{Multi-branch Deep Residual Structure.}
We construct relatively deeper residuals and apply residuals structure to all branches of our tree-based divergence network. In addition to the residuals within the residual group and the residuals from the input features to the end branches, we also add the residuals from intermediate branches to their own sub-branches as shown in Fig.~\ref{fig:pipeline}. Residual learning address the problem of long-term memory in deep SR network. Therefore, inspired by previous works~\cite{kim2016accurate,ledig2017photo,zhang2018image}, we utilize multi-branch deep residual structure to achieve more performance gain. We validate the effect of deep residual structure in Sec.~\ref{sec abl_study}.

\subsubsection{Divergence loss.}
Divergence loss used in divergence network, which composed of $L_2$ loss and modified triplet loss. The $L_2$ loss can be defined as 
\begin{equation}
    \small
    L_2^{D}=\sum ^{P}_{i=1}\left\| I_{D}^{i}-I_{HR}\right\| _{2}.
\end{equation}
It is worth noting that we used $L_2$ loss on all branches separately instead of just averaging them out with an overall $L_2$ loss. Because it allows for a more independent divergence of each branch, which is beneficial for the divergence stage. In order to make the divergence network produce divergence results more stable, we adopt triplet loss between all pairs of different predictions from divergence network. Our goal is to make the distance between $I_{D}^{i}$ and HR close and the distance between different members within $I_{D}$ farther. However, using triplet loss directly on RGB images causes the network to focus more on other differentiated directions (e.g., RGB space and luminance) than texture features. Therefore, we perform a series of processes on the triplet inputs shown in Fig.~\ref{fig:pipeline}. Firstly, we process $I_{D}^{i}$ by $G(\cdot)$:
\begin{equation}
\small
    G(I_{D}^{i})=\dfrac{Y_{D}^{i}-\mu_{Y_{D}^{i}}}{\sigma_{Y_{D}^{i}}},
\end{equation}
where $Y_{D}^{i}$ is Y channel of $I_{D}^{i}$ in YCbCr space, $\mu_{Y_{D}^{i}}$ and $\sigma_{Y_{D}^{i}}$ are mean and standard deviation of $Y_{D}^{i}$ respectively. $G(\cdot)$ operation enables network to focus more on the differentiation of texture space rather than color and luminance space. Secondly, the triplet inputs are converted to the residual domain where texture differences can be better represented. We express the residual of $I_{D}^{i}$ as 
\begin{equation}
\small
    res_{I_{D}^{i}}=\left | G(I_{D}^{i})- G(I_{HR}) \right |,
\end{equation}
where $\left |\cdot\right | $ is absolute value function. The pixel value of the residual images can oscillate between positive and negative, and absolute value function can reduce the checkerboard phenomenon caused by triplet loss as shown in Fig~\ref{abs_checkboard}. Overall, the formula for triplet loss is
\begin{equation}
\small
      trip(a,p,n)=Max[d(a,p)-d(a,n)+margin,0],
\end{equation}
where $a$, $p$ and $n$ are anchor, positive and negative, respectively. Therefore, we can represent our final triplet loss as
\begin{equation}
\small
    T_D=\frac{\sum ^{P}_{i=1} \sum ^{P}_{j=1,j\ne i}\beta_{ij} \ast trip(res_{I_{D}^{i}},zero,res_{I_{D}^{j}})}{(P(P-1))},
\label{eq:final_tp}
\end{equation}
where $zero$ is zero map and it is the positive in residual domain. $\beta $ is a attenuation coefficient which can be described as 
\begin{equation}
\small
\beta_{ij}=\theta ^{l-1}, l\in \left [ 1,L \right ],
\end{equation}
where $\theta$ is parameter and $\theta \in \left (0,1 \right ] $. The $l$ is index of the tree depths where the common parent branch of $I_{D}^{i}$ and $I_{D}^{j}$ is located. We use $\beta$ because that in a tree structure, differentiation should be progressive, i.e., branches with closer relatives should have relatively smaller degrees of divergence. Finally, our divergence loss is described as
\begin{equation}
\small
L_D=L_2^D + \alpha \ast T_D,
\end{equation}
where $\alpha$ is weight hyperparameter of $T_D$. Some detailed parameter settings will be introduced in Sec.~\ref{sec Dataset and Implementation Details}.

\subsection{Convergence Network}
\label{sec Convergence Network}
Combining the divergence results generated by the divergence network can produce more accurate results. We think that different areas on the predictions have different contribution weights for the final result. So we construct convergence network to merge divergent predictions weighted pixel-by-pixel, which can generate the result closer to the real HR image. Convergence network concatenates all the $P$ predictions of the divergence network and outputs weight map for each prediction $I_{D}^{i}$. Weight maps can be expressed as $W=F(Concat(I_{D});\Theta_{C})$, where $\Theta_{C}$ is the convergence network model parameter. We denote $W_i$ as the weight map of $I_{D}^{i}$. Every $I_{D}^{i}$ are element-wise multiplied by respective weight map $W_i$, and then all of the results are summed to produce the final SR result. Accordingly, SR result can be defined as
\begin{equation}
\small
    I_{SR}=\sum ^{P}_{i=1}\left( I_{D}^{i}\cdot W_{i}\right)\label{eq1},
\end{equation}
where $I_{SR}$ represents final SR result.

\subsubsection{Convergence loss.}
The goal of convergence network is to merge $I_{D}^{i}$ from divergence network. Therefore, loss function of convergence network is called convergence loss, which only consists of $L_2$ loss. We denote convergence loss as $ L_2^{C}$, which can be expressed as 
\begin{equation}
\small
    L_2^{C}=\left\| I_{SR}-I_{HR}\right\| _{2}.
\end{equation}
How to generate $I_{SR}$ has been introduced in Eq.~\ref{eq1}. The goal of convergence loss is to make the generated SR image get closer to the ground-truth HR image.

\subsection{Training Strategy} 
\label{sec Training Strategy}
The two networks in our framework are trained separately. We firstly train the divergence network into a stable status where it can generate super-resolution divergence predictions. We then freeze the parameters of divergence network and train the convergence network by enabling the whole pipeline. More details will be discussed in Sec.~\ref{sec Dataset and Implementation Details}.

\section{Experiment}
\subsection{Dataset and Implementation Details}
\label{sec Dataset and Implementation Details}
\subsubsection{D2CRealSR.} 
Existing RealSR datasets generally include x2, x3 and x4 scaling factor only, but lack of larger scaling factor. We collect a new dataset on x8 scaling factor, which called D2CRealSR. We construct the LR and HR pairs by zooming the lens of DSLR cameras. D2CRealSR consists of 115 image pairs and 15 image pairs are selected randomly for testing set; the rest pairs construct the training set. We use SIFT method to register the image pairs iterative: we first register the image pairs roughly and crop the main central area, then we align the brightness of the central area of image pairs and register them once again. After, we crop off the edges of aligned image pairs. The image size of each pair is 3,456$\times$2,304 after the alignment process.

\begin{table}[t]
 
\caption{Performance comparison on RealSR~\cite{cai2019toward}, DRealSR~\cite{wei2020component} and our proposed D2CRealSR datasets. The best results are \textbf{highlighted}. `-' indicates either the available model is not supported for such a test, or not open-sourced.}
\centering
\resizebox{1\linewidth}{!}
{
\begin{tabular}{ccccccccccccccc}
\toprule

                         & \multicolumn{6}{c}{RealSR}                                                                                      & \multicolumn{6}{c}{DRealSR (train on RealSR)}       & \multicolumn{2}{c}{D2CRealSR}                               \\ \cmidrule(lr){2-7}  \cmidrule(lr){8-13} \cmidrule(lr){14-15}
                         & \multicolumn{2}{c}{x2}                                     & \multicolumn{2}{c}{x3}                                      & \multicolumn{2}{c}{x4}                                     & \multicolumn{2}{c}{x2}                                     & \multicolumn{2}{c}{x3}                                     & \multicolumn{2}{c}{x4}                                     & \multicolumn{2}{c}{x8}                                     \\ \cmidrule(lr){2-3} \cmidrule(lr){4-5}
                         \cmidrule(lr){6-7}\cmidrule(lr){8-9}\cmidrule(lr){10-11}\cmidrule(lr){12-13}\cmidrule(lr){14-15}
\multirow{-4}{*}{Method} & \multicolumn{1}{c}{PSNR}    & \multicolumn{1}{c}{SSIM}    & \multicolumn{1}{c}{PSNR}     & \multicolumn{1}{c}{SSIM}    & \multicolumn{1}{c}{PSNR}    & \multicolumn{1}{c}{SSIM}                         & \multicolumn{1}{c}{PSNR}    & \multicolumn{1}{c}{SSIM}    & \multicolumn{1}{c}{PSNR}    & \multicolumn{1}{c}{SSIM}    & \multicolumn{1}{c}{PSNR}    & \multicolumn{1}{c}{SSIM}                         & \multicolumn{1}{c}{PSNR}    & \multicolumn{1}{c}{SSIM}                         \\ \midrule
Bicubic                  & 31.67                        & 0.887                        & 28.61                         & 0.810                        & 27.24                        & 0.764                        & 32.67                        & 0.877                        & 31.50                        & 0.835                        & 30.56                        & 0.820                        & 27.74                             &0.822                              \\
DRCN~\cite{kim2016deeply}                 & 33.42                        & 0.912                        & 30.36                         & 0.848                        & 28.56                        & 0.798                        & 32.46                        & 0.873
& 31.58                        & 0.838                        & 30.14                        & 0.816                        & 29.99                        & 0.833                        \\
SRResNet~\cite{ledig2017photo}                 & 33.17                        & 0.918                        & 30.65                         & 0.862                        & 28.99                        & 0.825                        & 32.85                        & 0.890 & 31.25                        & 0.841                        & 29.98                        & 0.822                        & 30.01                        & 0.864                        \\
EDSR~\cite{lim2017enhanced}                     & 33.88                        & 0.920                        & 30.86                         & 0.867                        & 29.09                        & 0.827                        & 32.86                        & 0.891                        & 31.20                        & 0.843 & 30.21                        & 0.817                        & 30.23                        & 0.868 \\
RCAN~\cite{zhang2018image}                     & 33.83                        & 0.923                        & 30.90                         & 0.864                        & 29.21                        & 0.824                        & 32.93 & 0.889                        & 31.76                        & 0.847 & 30.37                        & 0.825 &30.26 &0.868 \\
ESRGAN~\cite{wang2018esrgan}                   & 33.80                        & 0.922                        & 30.72                         & 0.866                        & 29.15                        & 0.826                        & 32.70                        & 0.889                        & 31.25                        & 0.842                        & 30.18                        & 0.821                        & 30.06        & 0.865        \\ 
SR-Flow~\cite{lugmayr2020srflow}                  & -                            & -                            & -                             & -                            & 24.20                         & 0.710                         & -                            & -                            & -                            & -                            & 24.97                              & 0.730                             & 23.11                        & 0.600                        \\ 
LP-KPN~\cite{cai2019toward}                   & 33.49                        & 0.917                        & 30.60                         & 0.865                        & 29.05                        & \textbf{0.834} & 32.77                        & -                            & 31.79 & -                            & 30.75 & -                            & -        & -      \\
CDC~\cite{wei2020component}                      & 33.96 & 0.925 & 30.99  & 0.869 & 29.24 & 0.827                        & 32.80                        & 0.888                        & 31.65                        & 0.847 & 30.41                        & \textbf{0.827} & 30.02                        & 0.841                        \\ 
\textbf{D2C-SR(Ours)}                     & \textbf{34.40} & \textbf{0.926} & \textbf{31.33} & \textbf{0.871} & \textbf{29.72} & {0.831} & \textbf{33.42} & \textbf{0.892} & \textbf{31.80} & \textbf{0.847} & \textbf{30.80} & {0.825} & \textbf{30.55} & \textbf{0.871} \\ \bottomrule
\end{tabular}
}
\label{tab:all_score}
 
\end{table}

\subsubsection{Existing datasets.} We also conduct experiments on existing real-world SR datasets: RealSR and DRealSR. RealSR has 559 scenes captured from DSLR cameras and align the image pairs strictly. 459 scenes for training and 100 scenes for testing. The image sizes of RealSR image pairs are in the range of 700\textasciitilde3000 and 600\textasciitilde3500. DRealSR has 83, 84 and 93 image pairs in testing set, 884, 783 and 840 image pairs in training set for x2, x3, x4 scaling factors, respectively. Note that we found some of the image pairs in DRealSR are misaligned, which is caused by the depth of field. So we only validate our results on testing set of DRealSR to show the performance of the cross-dataset of our method.

\subsubsection{Implementation details.} 
In the experiment, we set the number of branch layers as $L=2$, the number of child branches $C=2$. The basic branch module include $G=2$ residual groups and $B=4$ residual blocks in each group. We use Adam optimizer and set exponential decay rates as 0.9 and 0.999. The initial learning rate is set to $10^{-4}$ and then reduced to half every $2k$ epochs. For each training batch, we randomly extract 4 LR image patches with the size of $96\times96$. We implement D2C-SR method with the Pytorch framework and train the network using NVIDIA 2080Ti GPU.

\subsection{Comparisons with Existing Methods}
To evaluate our method, we train and test our model on our D2CRealSR with scaling factor x8 and an existing real-world SR dataset, RealSR with scaling factor x4, x3, x2. In addition, we validate on DRealSR tesing set for performance of cross-dataset. We compare our model with other state-of-the-art SISR methods, including DRCN~\cite{kim2016deeply}, SRResNet~\cite{ledig2017photo}, EDSR~\cite{lim2017enhanced}, RCAN~\cite{zhang2018image}, ESRGAN~\cite{wang2018esrgan}, LP-KPN~\cite{cai2019toward} and CDC~\cite{wei2020component}. The SISR results were evaluated on the Y channel in the YCbCr space using PSNR and SSIM. Among these SR methods, EDSR and RCAN are the classic SISR methods and DRCN is the open-source ensemble-based method. In addition, LP-KPN and CDC are designed to solve the real-world SR problem and they outperform on real-world SR benchmarks.

\subsubsection{Quantitative comparison.} 
The evaluation results of the SR methods, including our model and the other 8 methods are demonstrated in Table~\ref{tab:all_score}. The best results are highlighted and our method are 0.48$dB$ higher than the second ranked CDC method on x4 RealSR. Our model outperforms by a large margin on all the benchmarks and achieve state-of-the-art performance. 

\begin{figure}[t]
\begin{center}
\includegraphics[width=1\linewidth]{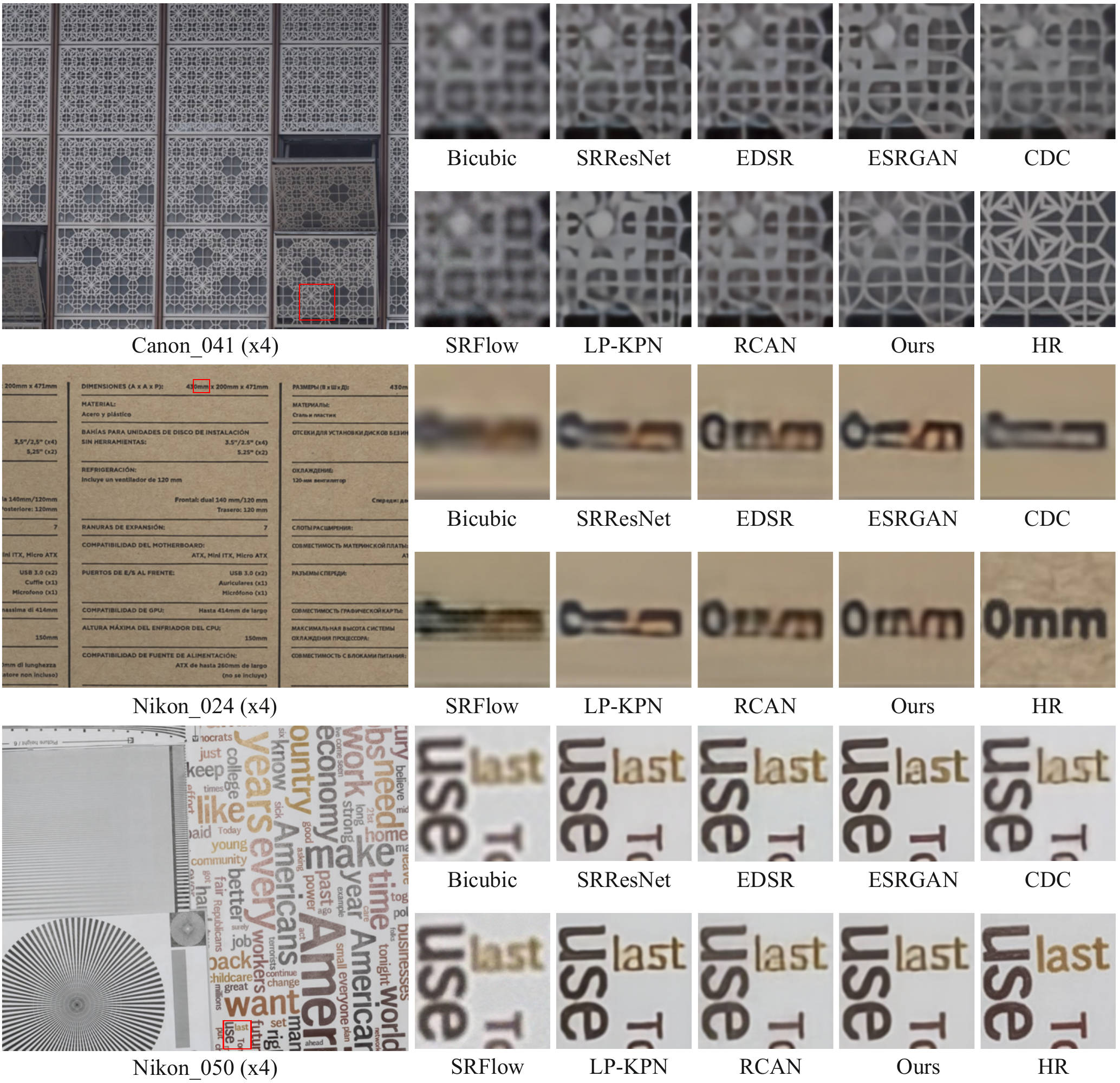}
\end{center}
 
 \caption{Comparison for x4 SR on RealSR~\cite{cai2019toward} dataset.}
\label{fig:Qua_fig}
 
\end{figure}

\begin{figure}[t]
\begin{center}
\includegraphics[width=1\linewidth]{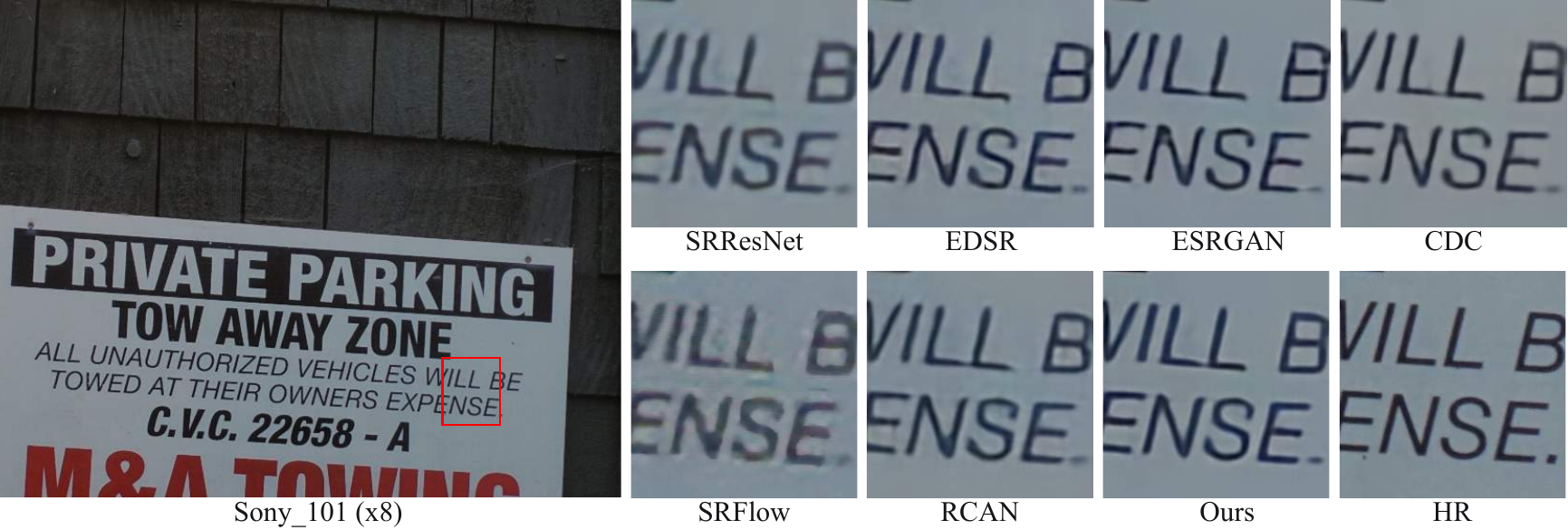}
\end{center}
 
  \caption{Comparison for x8 SR on our captured D2CRealSR dataset. }
\label{fig:Qua_fig_8x}
 
\end{figure}

\subsubsection{Qualitative comparison.} 
Visualization results of SR methods and ours on RealSR and D2CRealSR datasets are shown in Fig.~\ref{fig:Qua_fig} and Fig.~\ref{fig:Qua_fig_8x}. It is observed that existing SR methods (e.g., RCAN, LP-KPN, CDC) tend to restore the details to be thicker, blurry, or unnatural. Meanwhile, in Canon-041, Nikon-050 and Sony-101, our method recovered sharper and more accurate edge textures. In Nikon-024, we recover more information on the letters. From the visual comparisons, our method has the ability to recover richer and more accurate details.

\subsubsection{Visualization of D2C processes.}
\begin{figure}[t]
\begin{center}
\includegraphics[width=0.7\linewidth]{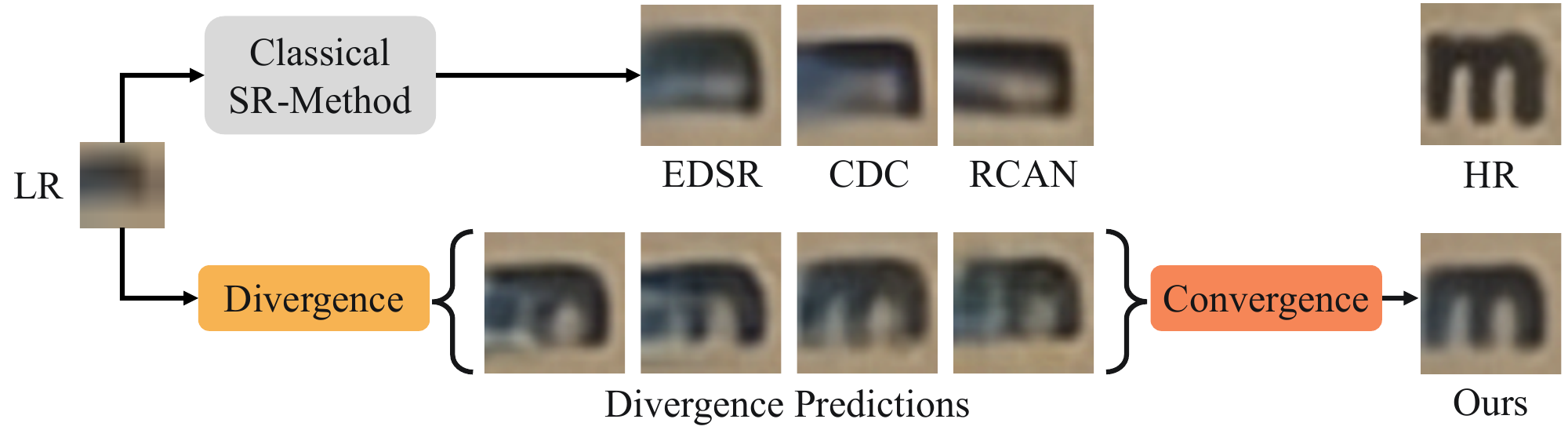}
\end{center}
 
  \caption{Visualization of D2C processes and other methods (x4).}
\label{fig:teaser}
\end{figure}

To further represent the progress of the D2C-SR, we visualize and compare the LR image, HR image, divergence intermediate results, and final results as well as the results of other comparison methods in Fig.~\ref{fig:teaser}. Classical SR methods directly use a single prediction to fit the distribution of all high-frequency details, thus obtain blurred outputs due to the ill-posed nature. Our two-stage approach explicitly learns the distribution of high-frequency details using a discrete manner, thus we can get rich and accurate texture details. 

\subsubsection{Comparison in perceptual metric.}
In recent years, the perceptual metric becomes another dimension to evaluate the image quality, and we also conduct a comparison on the perceptual metric in Table~\ref{tab:lpips}. Based on Sec.~\ref{sec Method}, we add vgg loss to train our model~\cite{ledig2017photo} to obtain a balance of multiple metrics. Experiments demonstrate that our method can achieve better performance on several evaluation measures, including the perceptual metric. 

\begin{table}[t]
 
\caption{Performance comparison in LPIPS on RealSR.}
\centering
\setlength\tabcolsep{6pt}
\resizebox{\linewidth}{!}
{
\begin{tabular}{cccccccccc}
\toprule
\multirow{2}{*}{Method} & \multicolumn{3}{c}{x2}                           & \multicolumn{3}{c}{x3}                           & \multicolumn{3}{c}{x4}                           \\ \cmidrule(lr){2-4}  \cmidrule(lr){5-7}  \cmidrule(lr){8-10}
                        & PSNR           & SSIM           & LPIPS          & PSNR           & SSIM           & LPIPS          & PSNR           & SSIM           & LPIPS          \\ \midrule
Bicubic                 & 31.67          & 0.887          & 0.223          & 28.61          & 0.810          & 0.389          & 27.24          & 0.764          & 0.476          \\ 
SRResNet~\cite{ledig2017photo}                & 33.17          & 0.918          & 0.158          & 30.65          & 0.862          & 0.228          & 28.99          & 0.825          & 0.281          \\ 
EDSR~\cite{lim2017enhanced}                    & 33.88          & 0.920          & 0.145          & 30.86          & 0.867          & 0.219          & 29.09          & 0.827          & 0.278          \\
RCAN~\cite{zhang2018image}                    & 33.83          & 0.923          & 0.147          & 30.90          & 0.864          & 0.225          & 29.21          & 0.824          & 0.287          \\
CDC~\cite{wei2020component}                  & 33.96          & 0.925          & 0.142          & 30.99          & 0.869          & 0.215          & 29.24          & 0.827          & 0.278          \\ 
\textbf{D2C-SR(Ours)}   & \textbf{34.39} & \textbf{0.926} & \textbf{0.136} & \textbf{31.31} & \textbf{0.870} & \textbf{0.214} & \textbf{29.67} & \textbf{0.830} & \textbf{0.270} \\ \bottomrule   
\end{tabular}
}
\label{tab:lpips}
 
\end{table}

\subsection{Applying the D2C Structure to Other Methods}
In Table~\ref{tab:d2c_others}, we apply the D2C structure to other methods and compare the performance with the original method. To make a fair comparison and reduce the influences caused by different sizes of parameters, we reduce the number of feature channels or the basic modules such that the number of parameters of the D2C counterpart model is smaller than the original model.  Our results show that multiple methods can achieve better performance with fewer number of parameters by applying the D2C structure.

\begin{table}[t]
 
\caption{Performance of applying our D2C structure to other methods on RealSR.}
\centering
\setlength\tabcolsep{6pt}
\resizebox{0.9\linewidth}{!}
{
\begin{tabular}{cccccccc}
\toprule
\multicolumn{1}{c}{\multirow{2}{*}{Method}} & \multicolumn{2}{c}{x2}                              & \multicolumn{2}{c}{x3}                              & \multicolumn{2}{c}{x4}                              & \multicolumn{1}{c}{\multirow{2}{*}{Parameters}} \\ 
\cmidrule(lr){2-3}  \cmidrule(lr){4-5}  \cmidrule(lr){6-7}
\multicolumn{1}{l}{}                  & \multicolumn{1}{c}{PSNR} & \multicolumn{1}{c}{SSIM} & \multicolumn{1}{c}{PSNR} & \multicolumn{1}{c}{SSIM} & \multicolumn{1}{c}{PSNR} & \multicolumn{1}{c}{SSIM} & \multicolumn{1}{l}{}                            \\ \midrule
SRResNet~\cite{ledig2017photo}                   & 33.17                    & 0.918                    & 30.65                    & 0.862                    & 28.99                    & \textbf{0.825}           & 1.52M                                           \\ 
\textbf{D2C-SRResNet}                 & \textbf{33.82}           & \textbf{0.920}           & \textbf{30.69}           & \textbf{0.862}           & \textbf{29.19}           & 0.822                    & \textbf{1.37M}                                  \\ \midrule
VDSR~\cite{kim2016accurate}                               & 31.39                    & 0.876                    & 30.03                    & 0.845                    & 27.07                    & 0.751                    & 0.67M                                           \\ 
\textbf{D2C-VDSR}                     & \textbf{34.08}           & \textbf{0.920}           & \textbf{30.29}           & \textbf{0.858}           & \textbf{28.21}           & \textbf{0.793}           & \textbf{0.67M}                                  \\ \midrule
RRDBNet~\cite{wang2018esrgan}                                & 33.44                    & 0.919                    & 30.29                    & 0.858                    & 28.34                    & 0.813                    & 16.7M                                           \\ 
\textbf{D2C-RRDBNet}                  & \textbf{33.49}           & \textbf{0.920}           & \textbf{30.33}           & \textbf{0.859}           & \textbf{28.47}           & \textbf{0.814}           & \textbf{9.16M}                                  \\ \midrule
RDN~\cite{zhang2018residual}                                   & 33.97                    & 0.922                    & 30.90                    & 0.864                    & 29.23                    & 0.824                    & 6.02M                                           \\ 
\textbf{D2C-RDN}                      & \textbf{34.03}           & \textbf{0.922}           & \textbf{30.93}           & \textbf{0.865}           & \textbf{29.32}           & \textbf{0.825}           & \textbf{5.59M}                                  \\ \midrule
EDSR~\cite{lim2017enhanced}                                   & 33.88                    & 0.920                    & 30.86                    & 0.867                    & 29.09                    & 0.827                    & 43.1M                                           \\ 
\textbf{D2C-EDSR}                     & \textbf{34.17}           & \textbf{0.924}           & \textbf{31.08}           & \textbf{0.868}           & \textbf{29.41}           & \textbf{0.829}           & \textbf{7.91M}                                  \\ \midrule
IMDN~\cite{hui2019lightweight}                                 & 33.59                    & 0.916                    & 30.74                    & 0.859                    & 29.17                    & 0.819                    & 0.874M                                          \\
\textbf{D2C-IMDN}                     & \textbf{33.95}           & \textbf{0.920}           & \textbf{30.98}           & \textbf{0.862}           & \textbf{29.45}           & \textbf{0.824}           & \textbf{0.738M}                                 \\ \bottomrule
\end{tabular}
}
\label{tab:d2c_others}
 
\end{table}

\subsection{Model Size Analyses}

We show comparisons about model size and performance in Fig.~\ref{fig:abl_model_size}. We list six sizes of models: 5.88M, 4.53M, 3.63M, 0.91M, 0.23M and 0.19M. These models are constructed by changing the number of residual groups $G$ and the number of residual block $B$. Our 0.23M model can achieve a better effect than other methods. At this PSNR level, CDC uses 39.92M, and RCAN also uses 15M parameters. Our baseline model chieve higher performance using 5.88M parameters only. Our method have a better trade-off between model size and performance.

\begin{flushleft}
\begin{minipage}[]{\textwidth}
 \begin{minipage}[]{0.48\textwidth}
\makeatletter\def\@captype{figure}\makeatother
       \includegraphics[width=\linewidth]{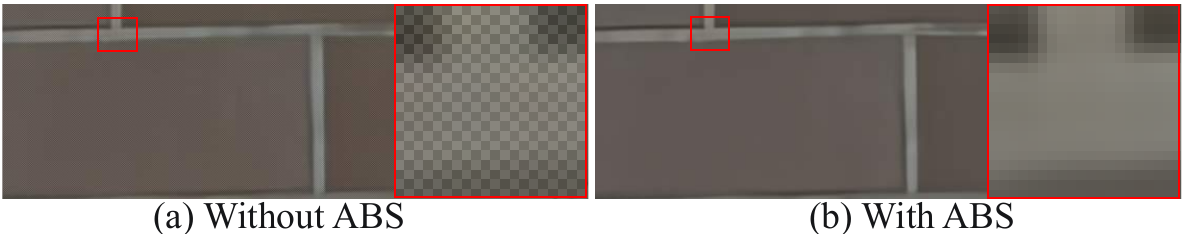}

     \caption{Absolute value function (ABS) reduces the checkerboard phenomenon.}
    \label{abs_checkboard}
    
     \makeatletter\def\@captype{table}\makeatother\caption{Effect study on the width and depth of the tree-based network (x4).}
      \label{tab:abl_tree_d}
      \setlength\tabcolsep{6pt}
     \resizebox{0.98\textwidth}{!}{
      \begin{tabular}{ccccc}
            \toprule
            \multicolumn{1}{c}{} & 1      & 2      & 3      & 4      \\ 
            \midrule
            Width ($C$)                 & 29.41  & 29.54 & 29.56 & \textbf{29.58} \\
            Depth ($L$)                & 29.30 & 29.54 & 29.63 & \textbf{29.64} \\
            \bottomrule
            \end{tabular}
     }

  \end{minipage}
  %
  \begin{minipage}[]{0.48\textwidth}
    \makeatletter\def\@captype{figure}\makeatother
        \includegraphics[width=0.95\linewidth]{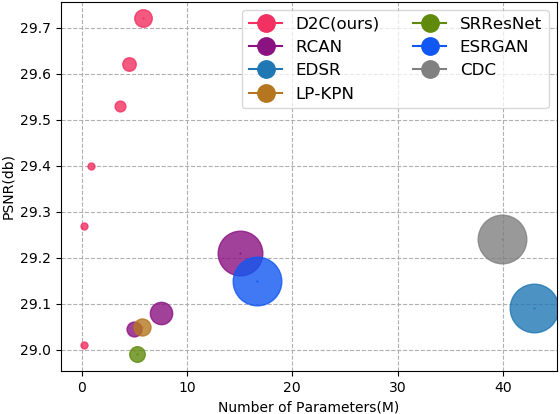}

     \caption{Performance vs. parameters.}
    \label{fig:abl_model_size}

  \end{minipage}

\end{minipage}
 
\end{flushleft}

\subsection{Ablation Studies}
\label{sec abl_study}
\subsubsection{The width and depth of tree-based network.}
Our divergence network is a tree-based architecture. We set different $L$ and $C$ to provide experimental evaluations on the width and depth of the tree. Because increasing the depth and width increases memory usage. So we do these experiments on a small baseline model, which has $G=2$ residual groups and each group has $B=1$ block. As show in Table~\ref{tab:abl_tree_d}, we increase the width and depth from 1 to 4 respectively on x4 scaling factor and show the changes in PSNR. By increasing the width or depth of the tree-based network, the performance of our D2C-SR improves.

\begin{table}[t]
 
 \setlength\tabcolsep{6pt}
\centering
     \caption{Effect of our divergence loss and multi-branch deep residual structure.}
    \label{tab:abl_summ}
     \resizebox{0.8\linewidth}{!}{
        \begin{tabular}{ccccccc}
\toprule
\multirow{2}{*}{}                  & \multicolumn{2}{c}{x2}           & \multicolumn{2}{c}{x3}           & \multicolumn{2}{c}{x4}                               \\ \cmidrule(lr){2-3} \cmidrule(lr){4-5} \cmidrule(lr){6-7} 
                                         & PSNR & SSIM  & PSNR & SSIM  & PSNR & SSIM \\ \midrule
\multicolumn{1}{l}{w/o. Divergence Loss} & 33.98                     & 0.920 & 31.12                     & 0.863 & 29.43                     & 0.824                     \\ 
\multicolumn{1}{l}{w/o. Multi Deep Res} & 34.28                    & 0.925 & 31.28                    & 0.870  & 29.68                    & 0.830                      \\ 
\multicolumn{1}{l}{D2C-SR(Ours)}           & \textbf{34.40}                      & \textbf{0.926} & \textbf{31.33}                     & \textbf{0.871} & \textbf{29.72}    & \textbf{0.831}                     \\ \bottomrule
\end{tabular}
     }
      
\end{table}

\begin{figure}[t]
 
\begin{center}
\includegraphics[width=1.0\linewidth]{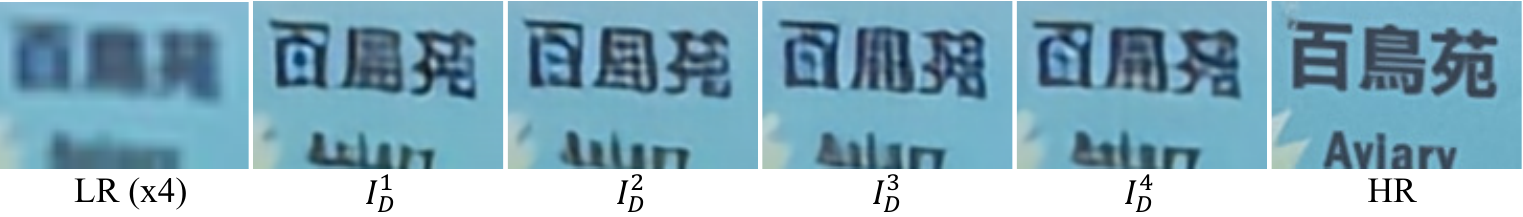}
\end{center}
 
    \caption{Visualization of the divergent predictions. }
\label{fig:triplet_demo}
 
\end{figure}

\begin{figure}[t]
 
\begin{center}
\includegraphics[width=0.8\linewidth]{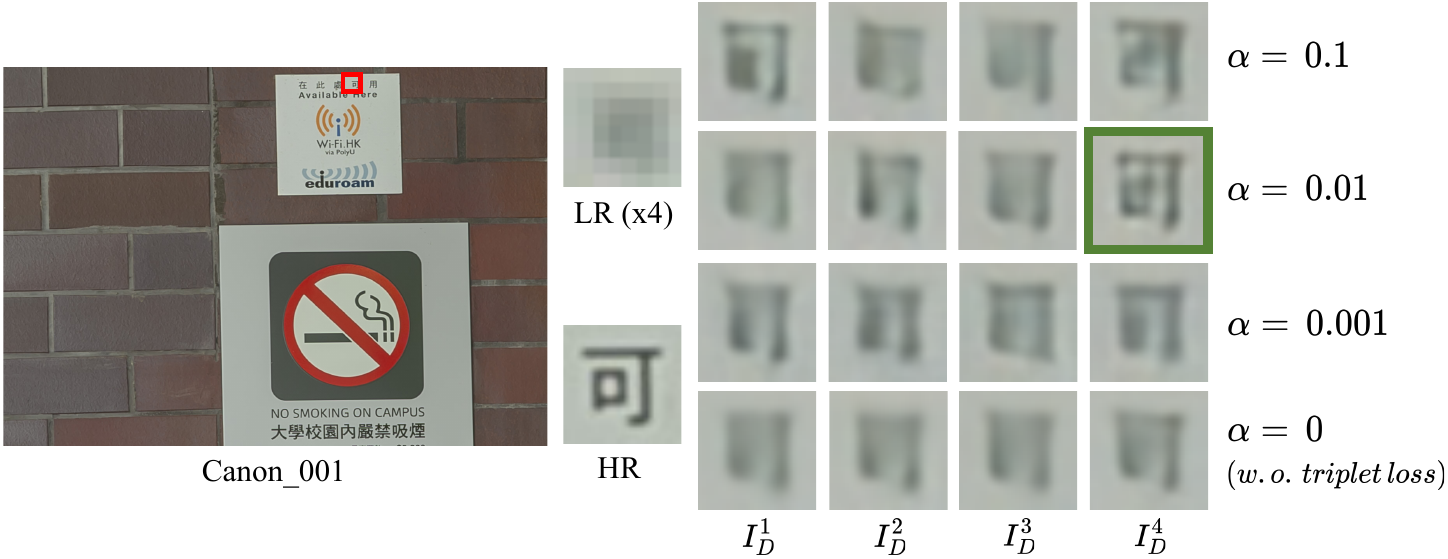}
\end{center}
 
   \caption{Comparisons on different weighting coefficient $\alpha$ in our divergence loss.}
\label{triplet_compare}
 
\end{figure}

\subsubsection{Divergence loss.} As mentioned in Sec.~\ref{sec Divergence Network}, the divergence loss is used to enforce the outputs from different branches divergent. In Fig.~\ref{fig:triplet_demo}, the visualization divergent results show that different branches produce different texture predictions. We verify the effectiveness of our divergence loss by removing the term of Eq.~\ref{eq:final_tp}. As shown in Table~\ref{tab:abl_summ} "w/o. divergence loss", the approach appears degraded. Because of the random nature of the model initialization, it is possible to lead to some degree of divergence in the results without using triplet loss and thus still gain some performance, it is still possible to gain some benefit from the effect of divergence. However, as mentioned in Sec.~\ref{sec:introduction}, it is not stable without the explicit divergence constraint, and thus may still bring degradation of performance. Further, we change the weight of the divergence loss and observe the changes that occur in different branches. As shown in Fig.~\ref{triplet_compare}, the results are relatively similar when divergence loss is not used. As the coefficients $\alpha$ increase, the differentiation of the different branches becomes more obvious. In this example, the results containing richer and more details appear. Theoretically, if $\alpha$ or the margin is too large it may lead to performance degradation, so we determine the $\alpha$ or the margin based on the model convergence. 

\subsubsection{Multi-branch deep residual structure.} As mentioned in Sec.~\ref{sec Divergence Network}, we also disable the structure to verify its effectiveness. As shown in Table~\ref{tab:abl_summ} “w/o. Multi Deep Res”, the PSNR decreased without the multi-branch deep residual structure. 

\begin{figure}[]
 
\begin{center}
\includegraphics[width=0.85\linewidth]{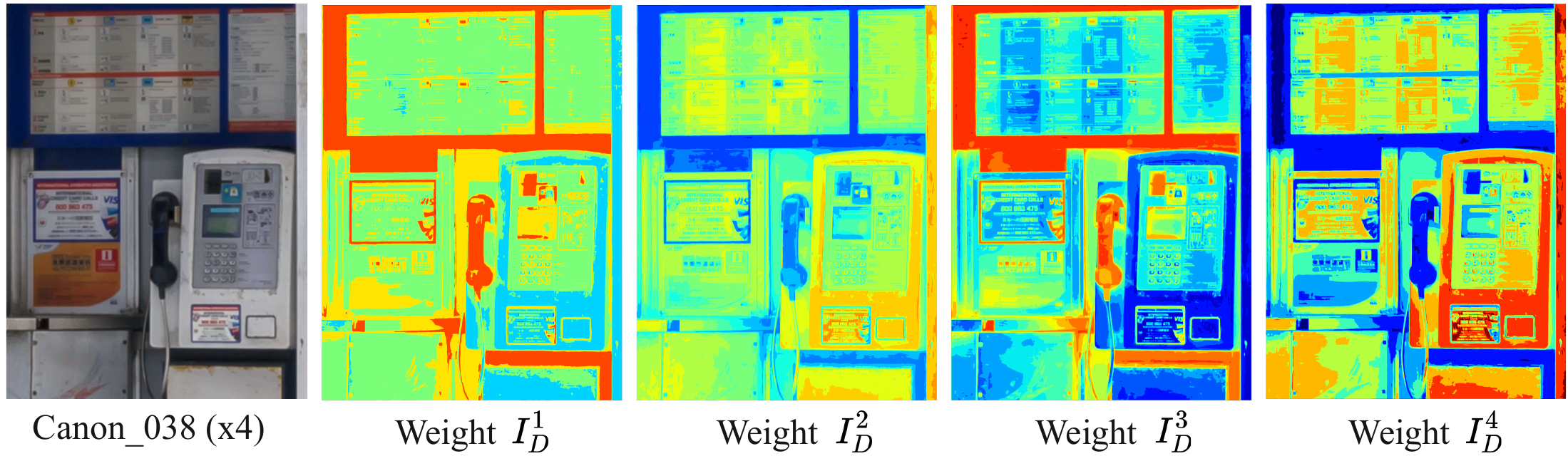}
\end{center}
 
    \caption{Weight maps in convergence stage. Red indicates higher values.}
\label{fig:weight_mask}
 
\end{figure}

\subsection{Visualization of Weight Maps} 
We assign spatial weights to the divergence results and use this weight for the fusion of the divergence results at the convergence stage. In Fig.~\ref{fig:weight_mask}, we visualize the weight maps using the heatmap. The visualization results show that predictions have some regional and texture correlation. We suggest that the weight generation depends mainly on the sharpness, edge style and texture preferences of the divergence results. We further measure the gap between the learned weight and the approximate ideal weight by finding the pixel that best matches the HR in the divergence results. The PSNR between the ideal fusion results and HR is $31.06$dB on RealSR x4, thus the gap is $1.34$dB ($29.72$dB ours).

\subsection{Simulated SISR and Real-World SISR} 
More works have focused on real-world SR because the great gap between simulated and real-world degradation hinders practical SR applications\cite{cai2019toward,wei2020component}. On the other hand, real-world degradation loses more information than bicubic compared with the HR, which also have been discussed in \cite{chen2019camera}, and ill-posed problem becomes more apparent in real-world datasets. Therefore, the real-world datasets can better reflect the effectiveness of our method.

\section{Conclusion}
In this study, we revisit the problem of image super-resolution and provide a new two-stage approach: divergence stage for multiple predictions learning as well as convergence for predictions fusion. Considering the limitations of traditional SR methods and SR space methods, we adopt a trade-off approach by using a finite discrete distribution to fit the high-frequency details. This allows the network to be more efficient and achieves state-of-the-art performance with much less computational cost. Futher, our D2C framework is a promising direction for image processing tasks like image inpainting as well as image denoising, and it is worth further exploration.

\noindent\textbf{Acknowledgment}
This work was supported by the National Natural Science Foundation of China (NSFC) under grants No. 61872067 and No. 61720106004.
\clearpage
%
%

\begin{thebibliography}{10}
\providecommand{\url}[1]{\texttt{#1}}
\providecommand{\urlprefix}{URL }
\providecommand{\doi}[1]{https://doi.org/#1}

\bibitem{anwar2020deep}
Anwar, S., Khan, S., Barnes, N.: A deep journey into super-resolution: A
  survey. ACM Computing Surveys (CSUR)  \textbf{53}(3),  1--34 (2020)

\bibitem{bahat2020explorable}
Bahat, Y., Michaeli, T.: Explorable super resolution. In: CVPR. pp. 2716--2725
  (2020)

\bibitem{buhler2020deepsee}
Buhler, M.C., Romero, A., Timofte, R.: Deepsee: deep disentangled semantic
  explorative extreme super-resolution. In: ACCV (2020)

\bibitem{cai2019toward}
Cai, J., Zeng, H., Yong, H., Cao, Z., Zhang, L.: Toward real-world single image
  super-resolution: A new benchmark and a new model. In: CVPR. pp. 3086--3095
  (2019)

\bibitem{chen2019camera}
Chen, C., Xiong, Z., Tian, X., Zha, Z.J., Wu, F.: Camera lens super-resolution.
  In: CVPR. pp. 1652--1660 (2019)

\bibitem{dai2015jointly}
Dai, D., Timofte, R., Van~Gool, L.: Jointly optimized regressors for image
  super-resolution. In: Computer Graphics Forum. vol.~34, pp. 95--104 (2015)

\bibitem{goodfellow2014generative}
Goodfellow, I.J., Pouget-Abadie, J., Mirza, M., Xu, B., Warde-Farley, D.,
  Ozair, S., Courville, A., Bengio, Y.: Generative adversarial networks. arXiv
  preprint arXiv:1406.2661  (2014)

\bibitem{gu2019blind}
Gu, J., Lu, H., Zuo, W., Dong, C.: Blind super-resolution with iterative kernel
  correction. In: CVPR. pp. 1604--1613 (2019)

\bibitem{hui2019lightweight}
Hui, Z., Gao, X., Yang, Y., Wang, X.: Lightweight image super-resolution with
  information multi-distillation network. In: Proceedings of the 27th acm
  international conference on multimedia. pp. 2024--2032 (2019)

\bibitem{kim2016accurate}
Kim, J., Kwon~Lee, J., Mu~Lee, K.: Accurate image super-resolution using very
  deep convolutional networks. In: CVPR. pp. 1646--1654 (2016)

\bibitem{kim2016deeply}
Kim, J., Lee, J.K., Lee, K.M.: Deeply-recursive convolutional network for image
  super-resolution. In: CVPR. pp. 1637--1645 (2016)

\bibitem{ledig2017photo}
Ledig, C., Theis, L., Husz{\'a}r, F., Caballero, J., Cunningham, A., Acosta,
  A., Aitken, A., Tejani, A., Totz, J., Wang, Z., et~al.: Photo-realistic
  single image super-resolution using a generative adversarial network. In:
  CVPR. pp. 4681--4690 (2017)

\bibitem{lim2017enhanced}
Lim, B., Son, S., Kim, H., Nah, S., Mu~Lee, K.: Enhanced deep residual networks
  for single image super-resolution. In: CVPRW. pp. 136--144 (2017)

\bibitem{lugmayr2020srflow}
Lugmayr, A., Danelljan, M., Van~Gool, L., Timofte, R.: Srflow: Learning the
  super-resolution space with normalizing flow. In: ECCV. pp. 715--732 (2020)

\bibitem{michelini2019multigrid}
Michelini, P.N., Liu, H., Zhu, D.: Multigrid backprojection super--resolution
  and deep filter visualization. In: AAAI. vol.~33, pp. 4642--4650 (2019)

\bibitem{pan2020real}
Pan, Z., Li, B., Xi, T., Fan, Y., Zhang, G., Liu, J., Han, J., Ding, E.: Real
  image super resolution via heterogeneous model ensemble using gp-nas. In:
  ECCV. pp. 423--436. Springer (2020)

\bibitem{sajjadi2017enhancenet}
Sajjadi, M.S., Scholkopf, B., Hirsch, M.: Enhancenet: Single image
  super-resolution through automated texture synthesis. In: ICCV. pp.
  4491--4500 (2017)

\bibitem{shi2016real}
Shi, W., Caballero, J., Husz{\'a}r, F., Totz, J., Aitken, A.P., Bishop, R.,
  Rueckert, D., Wang, Z.: Real-time single image and video super-resolution
  using an efficient sub-pixel convolutional neural network. In: CVPR. pp.
  1874--1883 (2016)

\bibitem{simonyan2014very}
Simonyan, K., Zisserman, A.: Very deep convolutional networks for large-scale
  image recognition. arXiv preprint arXiv:1409.1556  (2014)

\bibitem{song2020efficient}
Song, D., Xu, C., Jia, X., Chen, Y., Xu, C., Wang, Y.: Efficient residual dense
  block search for image super-resolution. In: AAAI. vol.~34, pp. 12007--12014
  (2020)

\bibitem{sun2012super}
Sun, L., Hays, J.: Super-resolution from internet-scale scene matching. In:
  {Proc. ICCP}. pp. 1--12 (2012)

\bibitem{timofte2013anchored}
Timofte, R., De~Smet, V., Van~Gool, L.: Anchored neighborhood regression for
  fast example-based super-resolution. In: ICCV. pp. 1920--1927 (2013)

\bibitem{timofte2014a+}
Timofte, R., De~Smet, V., Van~Gool, L.: A+: Adjusted anchored neighborhood
  regression for fast super-resolution. In: ACCV. pp. 111--126 (2014)

\bibitem{wang2017ensemble}
Wang, L., Huang, Z., Gong, Y., Pan, C.: Ensemble based deep networks for image
  super-resolution. Pattern recognition  \textbf{68},  191--198 (2017)

\bibitem{wang2018esrgan}
Wang, X., Yu, K., Wu, S., Gu, J., Liu, Y., Dong, C., Qiao, Y., Change~Loy, C.:
  Esrgan: Enhanced super-resolution generative adversarial networks. In: ECCV.
  pp.~0--0 (2018)

\bibitem{wang2020deep}
Wang, Z., Chen, J., Hoi, S.C.: Deep learning for image super-resolution: A
  survey. {IEEE Trans. on Pattern Analysis and Machine Intelligence}  (2020)

\bibitem{wei2020component}
Wei, P., Xie, Z., Lu, H., Zhan, Z., Ye, Q., Zuo, W., Lin, L.: Component
  divide-and-conquer for real-world image super-resolution. In: ECCV. pp.
  101--117 (2020)

\bibitem{xiong2018gradient}
Xiong, D., Gui, Q., Hou, W., Ding, M.: Gradient boosting for single image
  super-resolution. Information Sciences  \textbf{454},  328--343 (2018)

\bibitem{yang2013fast}
Yang, C.Y., Yang, M.H.: Fast direct super-resolution by simple functions. In:
  ICCV. pp. 561--568 (2013)

\bibitem{yang2008image}
Yang, J., Wright, J., Huang, T., Ma, Y.: Image super-resolution as sparse
  representation of raw image patches. In: CVPR. pp.~1--8 (2008)

\bibitem{zhang2019zoom}
Zhang, X., Chen, Q., Ng, R., Koltun, V.: Zoom to learn, learn to zoom. In:
  CVPR. pp. 3762--3770 (2019)

\bibitem{zhang2018image}
Zhang, Y., Li, K., Li, K., Wang, L., Zhong, B., Fu, Y.: Image super-resolution
  using very deep residual channel attention networks. In: ECCV. pp. 286--301
  (2018)

\bibitem{zhang2018residual}
Zhang, Y., Tian, Y., Kong, Y., Zhong, B., Fu, Y.: Residual dense network for
  image super-resolution. In: CVPR. pp. 2472--2481 (2018)

\end{thebibliography}

\end{document}